\documentclass[manuscript,nonacm]{acmart}
\AtBeginDocument{%
  \providecommand\BibTeX{{%
    \normalfont B\kern-0.5em{\scshape i\kern-0.25em b}\kern-0.8em\TeX}}}

\begin{document}

\title{Doubting AI Predictions: Influence-Driven Second Opinion Recommendation}

\author{Maria De-Arteaga}
\email{dearteaga@mccombs.utexas.edu}
%\email{webmaster@marysville-ohio.com}
\affiliation{%
  \institution{University of Texas at Austin}
  %\streetaddress{P.O. Box 1212}
  \city{Austin}
  \state{TX}
  \country{USA}
  \postcode{78712}
}

\author{Alexandra Chouldechova}
\affiliation{%
  \institution{Carnegie Mellon University}
 % \streetaddress{1 Th{\o}rv{\"a}ld Circle}
  \city{Pittsburgh}
  \country{USA}}
%\email{larst@affiliation.org}

\author{Artur Dubrawski}
\affiliation{%
  \institution{Carnegie Mellon University}
  \city{Pittsburgh}
  \country{USA}
}

\renewcommand{\shortauthors}{De-Arteaga, et al.}

\begin{abstract}
Effective human-AI collaboration requires a system design that provides humans with meaningful ways to make sense of and critically evaluate algorithmic recommendations. In this paper, we propose a way to augment human-AI collaboration by building on a common organizational practice: identifying experts who are likely to provide complementary opinions. When machine learning algorithms are trained to predict human-generated assessments, experts' rich multitude of perspectives is frequently lost in monolithic algorithmic recommendations. The proposed approach aims to leverage productive disagreement by (1) identifying whether some experts are likely to disagree with an algorithmic assessment and, if so, (2) recommend an expert to request a second opinion from. 
\end{abstract}

\maketitle
\section{Introduction}

When machine learning is used with the goal of improving experts' decisions, it is often the case that an AI system makes recommendations, but the ultimate decision is made by a human expert. This setup is common in healthcare, human resources, public services, and the criminal justice system. In such contexts of algorithmic decision support, accurate predictions are often not enough. For both the human and the machine to add value to a human-AI team, the human must have effective means at its disposal to make sense of algorithmic recommendations. Without this, algorithm aversion~\citep{dietvorst2015algorithm}, automation bias~\citep{skitka2000accountability}, or other forms of detrimental integration of AI recommendations into decisions may hamper the usefulness of AI predictions. As evidenced by an in-depth field study investigating how AI tools are used by diagnostic radiologists, physicians' professional and legal responsibility for every one of their decisions drive them to look for ways to interrogate AI recommendations, often unsuccessfully~\citep{lebovitz2020incorporate}. Thus, effective human-AI collaboration remains an elusive goal in many settings. While model interpretability and explainations have been proposed as a way to enable humans to make sense of AI recommendations and to critically integrate them into decisions (e.g.~\citet{caruana2015intelligible, ribeiro2016should}), it has also been shown that explanations may lead to over-reliance~\citep{lakkaraju2017selective}, and may fail to improve the quality of decisions~\citep{bansal2021does}. Alternatively, the importance of communicating model uncertainty has been increasingly emphasized as a key feature of algorithmic transparency~\citep{bhatt2020uncertainty, mcgrath2020does, bansal2021does}, but this piece of information alone may often not be enough to resolve ambiguity and improve the quality of decisions. 

Rather than letting the algorithm speak for (and against) itself, we build on current organizational practices of relying on second opinions provided by other experts. Previous work has explored the use of machine learning to determine \textit{when} to ask for a second opinion~\citep{raghu2018direct}. This paper tackles the question of \textit{who} to ask for a second opinion, while also addressing the question of when to do it. 

We propose two variants of the methodology, both of which assume access to historical experts' assessments. The first approach builds separate predictive models for each expert and chooses who to ask for a second opinion based on the prediction of these models and their comparison to the recommendation of an AI tool being used for decision support. The second approach leverages influence functions to estimate the influence of individual experts over the predictions of a single model trained to predict experts' decisions. In both variants, the magnitude of the predicted probabilities and the influence, respectively, can also serve as an indication of whether and when to ask. In cases where the AI tool being used for decision support is trained using labels that correspond to human assessments (e.g. radiologists), the latter has the advantage of being directly linked to the AI recommendation, by measuring experts' influence over it. Naturally, this approach is not only useful in identifying who is likely to disagree with an opinion, and can also be used to identify experts who can argue \textit{in favor} of the AI recommendation, by choosing the expert who most positively influences its prediction. 

\section{Related Work}

Human-AI collaboration is a broad space that considers different forms of human-AI teamwork. One type of collaboration considers ``division of labor" approaches, in which some instances are routed to a human and some instances are routed to an algorithm \citep{madras2018predict, wilder2020learning, gao2021human}, with the core idea being that the algorithm can specialize on the instances that are particularly hard for humans to assess. These approaches consider that the algorithm may be the final decision-maker for some instances, in contrast to high-stakes settings where algorithmic recommendations are provided to a human who makes the ultimate decision. In the context of human-in-the-loop frameworks, researchers have studied over- and under-reliance on algorithms~\citep{dietvorst2015algorithm, de2020case, buccinca2021trust}, the role of explanations~\citep{zhang2020effect, bansal2021does} and the importance of backward compatibility \citep{bansal2019case}.

The crucial role of uncertainty in humans-in-the-loop settings has been emphasized \citep{bhatt2020uncertainty, mcgrath2020does}, and it has motivated research on novel statistical methodologies as well as human-computer interaction. Recent work has also explored ways of estimating uncertainty in deep learning~\citep{gal2016dropout, maddox2019simple}. Another line of work studies the impact that communicating model uncertainty may have on human decisions and algorithmic reliance~\citep{zhang2020effect, bansal2021does}. 

The use of machine learning to decide \textit{when} to ask for a second opinion has been explored by~\citep{raghu2018direct}. In the context studied by~\citet{raghu2018direct} humans are always responsible for making assessments/predictions, and machine learning is used to estimate uncertainty in experts' decisions, in order to determine which cases are most likely to benefit from a second opinion. In this work we focus on the problem of \textit{who} to ask for a second opinion, epsecially considering cases in which the first opinion is provided by an algorithm.

A core element of our methodology relies on influence functions. The local influence method~\citep{cook1986assessment} is an approach from robust statistics that estimates the effect of minor perturbations of a model over a functional, such as the loss or the predicted probability. In machine learning, it has been used as a means to explain complex models and as a way to generate adversarial attacks~\citep{koh2017understanding}. Most recently, the local influence method has also been used to estimate the influence of individual experts over predictions of a model trained on human assessments, with the goal of bridging the gap between an algorithm's and experts' target objectives ~\citep{de2021leveraging}.

\section{Methodology}
\label{meth}

\subsection{Problem Formulation}

We assume a standard supervised learning set up, with features $x \in \mathcal{X}$ and labels $y \in \mathcal{Y}$. We also assume there is a set of experts $\{h_1, h_2, ...h_k \}$, for whom historical assessments $d_{h_i}(x) \in \mathcal{D}$ for $x \in \mathcal{X}$ are available. We note that the specific instances $x \in \mathcal{X}$ for which we have human assessments $\mathcal{D}$ and labels $\mathcal{Y}$ need not be the same, nor do we need to have assessments from every expert for each instance. For example, in some cases there may only be one expert assessment available per instance. 

Assume that an AI tool used for decision support provides a prediction $\hat{y}(x)$. The task is to identify $h_{ask}(x) \in \{h_1, h_2, ...h_k \}$ such that $d_{h_i}(x)\neq \hat{y}(x)$.
In many domains, the labels $y \in \mathcal{Y}$ used to train the model correspond to human assessments $d \in \mathcal{D}$. Among other domains, this is frequent in some healthcare diagnostic applications, such as radiology. While the proposed approaches are not constrained to this assumption, they do have additional benefits in such settings, given the tight connection between the experts' assessments and the AI tool. For this reason, and to aid in clarity, we assume for the remainder of the paper that the AI tool is trained to predict historical human assessments, $D \subseteq \mathcal{D}$.

Let $\hat{f}_D$ denote a predictive model of expert decisions, $\hat{f}_D=\hat{P}(D=1 | X)$, which does not differentiate between who provided each label, as is common when training predictive models using human assessments as labels. We assume this to be the AI tool used for decision support, which yields binary predictions $\hat{d}$ based on a threshold $\tau$ (in the general case, this predictions correspond to $\hat{y}$).  
Given a new case $x \in \mathcal{X}$, the task is to identify an expert to ask a second opinion from, $h_{ask}(x) \in \{h_1, h_2, ...h_k \}$, such that $d_{h_{ask}}(x)\neq \hat{d}(x)$.

\subsection{Proposed Approaches}

\textit{Independent models.} The first proposed approach relies on training individual models for each expert, and corresponds to the naive approach of using machine learning to identify sources of second opinion. While simple, this formulation has the advantage of potentially being able to capture heterogeneity across experts' decisions without requiring an increase in the complexity of the algorithms used to model experts, and may be a good fit when there is enough data generated by each expert.

For each expert $h_i \in \{h_1, h_2, ...h_k \}$, train a (calibrated) model to predict its assessments, $\hat{f}_{h_i}$. The provider of the second opinion, $h_{ask}$ can be selected as:

\begin{equation}
  h_{ask}(x)  = \left\{ \begin{array}{ll}
\mbox{argmin}_i(\hat{f}_{h_i} (x)) \mbox{ if } \hat{d}(x) = 1 \\
\mbox{argmax}_i(\hat{f}_{h_i} (x))  \mbox{ if } \hat{d}(x) = 0 \\
\end{array}\right. 
\label{eq:second_opinion_ind}
\end{equation}

The predicted probabilities $\hat{f}_{h_i} (x)$ can also be used as an indication of whether someone is likely to disagree, yielding a formulation that also considers when to ask for a second opinion, thus opening the possibility for $h_{ask}(x)$ to be empty,

\begin{equation}
  h_{ask}(x)  = \left\{ \begin{array}{ll}
\mbox{argmin}_i(\{\hat{f}_{h_i} (x) : \hat{f}_{h_i} (x)< \tau \}) \mbox{ if } \hat{d}(x) = 1 \\
\mbox{argmax}_i(\{\hat{f}_{h_i} (x) : \hat{f}_{h_i} (x)> \tau \})  \mbox{ if } \hat{d}(x) = 0 \\
\end{array}\right. 
\label{eq:second_opinion_ind_empty}
\end{equation}

\textit{Influence-driven selection.} The second proposed approach does not require training of separate models for each expert, which has the advantage of not requiring a large amount of data per expert. Additionally, its choice of second opinion is intricately link to the prediction provided by the AI tool, and to whose decisions would most influence that prediction in a different direction. 

Assume the training data $X$ has dimensions $m\times n$, and each instance $x$ has been labeled by an expert $h$. Which expert labeled each case is stored in a vector $a \in  \mathbb{R}^{m\times 1}$, such that each entry of the vector, $a^j$, denotes the expert who labeled the instance $x^j$, $a^j \in \{h_1, h_2, ..., h_k\}$. Following the local influence method~\citep{cook1986assessment}, let $w_{h_i} \in \mathbb{R}^{m\times 1}$ be a perturbation of the training data that marginally up-weights the importance given to decision-maker $h_i$, defined as follows, where $w_{h}^j$ denotes the $j$th entry of the vector $w_{h}$:

\begin{equation}
\begin{array}{rcl}
w_{h_i}^j & = & \left\{ \begin{array}{rcl}
1 + \varepsilon & \mbox{for} & a^j == h_i \\
1 & \mbox{for} & a^j \neq h_i  \\
\end{array}\right. ,
\end{array}
\label{eq:w}
\end{equation}

The influence of this perturbation over the predicted probability indicates how would the predicted probability change if the training data was perturbed in the direction corresponding to an expert $h$. This influence can be defined as follows:

\begin{equation}
\begin{array}{ll}
 \mathcal{I}_{up,f_D}(w_h,x_{test}) &  :=  \left. \frac{\partial P(y_{test}|x_{test},\hat{\theta}_{w_h})} { \partial \epsilon}\right \rvert_{\epsilon=0}  \\ & = \left.  \nabla_{\theta}P(y_{test}|x_{test},\hat{\theta}_{w_h})^T \frac{\partial \hat{\theta}_{w_h}}{\partial \epsilon} \right \rvert_{\epsilon=0}\\ [10pt]
\end{array}
  \label{eq:infl_p}
\end{equation}

Once this influence is estimated, which can be done following the approach in~\citet{de2021leveraging}, it can be used to select $h_{ask}$, based on whose influence would most influence the prediction in a direction opposite to that of the provided recommendation,

\begin{equation}
  h_{ask}(x)  = \left\{ \begin{array}{ll}
\mbox{argmin}_i(\mathcal{I}_{up,f_D}(w_{h_i},x))& \mbox{if } \hat{d}(x) =1 \\
\mbox{argmax}_i(\mathcal{I}_{up,f_D}(w_{h_i},x)) & \mbox{if } \hat{d}(x) = 0 \\
\end{array}\right. 
\label{eq:second_opinion_inf}
\end{equation}

Naturally, the choice to request a second opinion may also be informed by whether any expert has an influence in the direction opposite to $\hat{f}_D(x)$, and even further constrained by choosing a threshold for the minimum magnitude of influence required to request an opinion. Making use of the fact that the influence itself is informative for knowing which cases may be expected to have differing views and conflicting opinions, the choice can be reformulated as,

\begin{equation}
  h_{ask}(x)  = \left\{ \begin{array}{l}
\mbox{argmin}_i(\{\mathcal{I}_{h_i}(x)  :  \mathcal{I}_{h_i}(x)<0\}) \mbox{ if } \hat{d}(x) = 1 \\
\mbox{argmax}_i(\{\mathcal{I}_{h_i}(x) : \mathcal{I}_{h_i}(x)>0\}) \mbox{ if } \hat{d}(x) = 0  \\
\end{array}\right. 
\label{eq:second_opinion_inf_empty}
\end{equation}

where $\mathcal{I}_{h_i}(x)$ denotes $\mathcal{I}_{up,f_D}(w_{h_i},x)$ for simplification in the notation, and $h_{ask}$ may be empty if nobody is likely to provide a differing opinion.

\section{Experiments}

\subsection{Results}

\begin{figure*}[h]
    \centering
%     \begin{subfigure}
% \centering
\includegraphics[width=.45\textwidth]{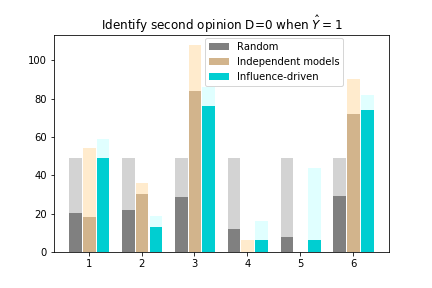}
%\rule{6cm}{1cm}% in place of the image
\def\baselinestretch{1}
%\end{subfigure}
%\hspace{0.1in}
% \begin{subfigure}
% \centering
\includegraphics[width=.45\textwidth]{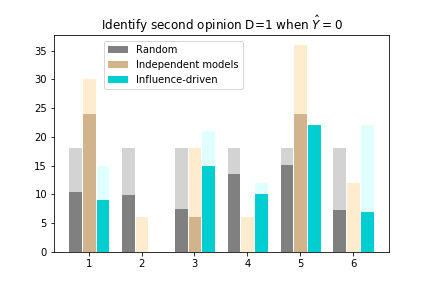}
%\def\baselinestretch{1}
%\end{subfigure}
    \caption{Performance of proposed approaches to recommend who to ask for a second opinion, assessing whether and to what extent the methods can correctly retrieve disagreement when it exists. Each bar indicates frequency with which a decision maker (1-6) is selected for a second opinion, with shaded portion indicating frequency with which the experts' assessment $d_i$ matches the second opinion sought.}
    \label{fig:res}
\end{figure*}

We conduct experiments using a data set containing subjective quality assessment of digital colposcopies~\citep{fernandes2017transfer}, which records the independent assessment of six physicians for each case. We use the dataset corresponding to the use of a green filter for feature extraction given its higher degree of disagreement among experts. There is a total of 588 individual assessments, which correspond to 98 unique cases, $66\%$ of which do not have complete agreement across experts.

We first pre-process the data by applying PCA in order to reduce feature colinearity, as this would preclude the reliable estimation of the inverse of the Hessian when estimating influence functions. To estimate $\hat{f}_D$ we train a single logistic regression model that jointly predicts physicians' assessment and estimate each expert's influence as described in Section~\ref{meth},  using 3-fold cross-validation to obtain recommendations for second opnion in the entire data set. Separately, we train six independent logistic regression models (one per physician) and use 3-fold cross-validation to estimate predictions for each $\hat{f}_{h_i}$ in the entire data set.

Figure~\ref{fig:res} shows the performance of the two approaches described in Section~\ref{meth}. For each, we assess whether and to what extent they can correctly retrieve a second opinion that disagrees with the AI tool's prediction, whenever disagreement exists. That is, we consider the subset of cases without full agreement, and assess if given a prediction of the AI tool, $\hat{d}(x)$, the methods can correctly identify someone with a differing opinion, $h_{ask}$, such that $d_{h_{ask}}(x)\neq \hat{d}(x)$. For each expert, Figure~\ref{fig:res} shows the frequency with which the expert is chosen and the frequency with which this choice is correct. For comparison, the results of random selection are also shown, where the shown frequency of being correct corresponds to the base rate with which each expert makes each decision. Table~\ref{tab:rates} shows the aggregate accuracy rate for each of the two proposed models.       
\begin{table}[h]
    \centering{
    \begin{tabular}{lrrr}\hline
        Method & Overall & ($\hat{Y}=1$)&  ($\hat{Y}=0$)\\ \hline
        Indep. models & 0.64 & 0.69 & 0.5 \\ %\hline
        Influence-driven & 0.72 & 0.73 & 0.69\\ \hline
    \end{tabular}}
    \caption{Accuracy of correctly identifying a second opinion that will disagree with the AI tool. Both overall performance and performance disaggregated by models' prediction (and thus by the type of opinion sought) are reported. }
    \label{tab:rates}
\end{table}

\subsection{Analysis}
 Across experts, the influence-driven approach almost always outperforms both the independent model approach and the baseline comparison, as shown in Figure~\ref{fig:res}. This is also true when considering performance conditioned on the AI tool's predictions, as shown in both Figure~\ref{fig:res} and Table~\ref{tab:rates}. Here, we consider separately the set of cases where $\hat{d}=1$ and thus we seek an opinion $d_{h_{ask}}=0$, and the cases where $\hat{d}=0$ and thus we seek an opinion $d_{h_{ask}}=1$. The gains in performance are particularly large for the set $\hat{d}=0$, which is the less prevalent assessment. 

In addition to raw performance metrics, other properties of the approaches may impact their usability in practice. In particular, it is worth noting that the influence-driven approach distributes the choice of second opinion requests more smoothly across experts. This has important implications for its usability, as it would be likely unfeasable and undesirable to overburden a single expert with requests. Furthermore, this may also translate into benefits for the individuals who are subjected to the algorithm. Different experts will have different areas of expertise, and some may display different performance and biases across subpopulations. Thus, overrelying on a single expert could disadvantage cases that would be better served by other experts.

\section{Conclusion and Future Work}

This paper proposes the use of machine learning for second opinion recommendation as a form to augment human-AI collaboration. Grounded on organizational and domain-specific practices of requesting and seeking alternative perspectives during the decision-making process, the proposed approach seeks to support experts by identifying who is most likely to provide alternative points of view. 

Future work will seek to further validate the proposed approaches across datasets, assessing the model's ability to correctly identify experts who can provide a differing second opinion. An important component of these experiments will be to include larger datasets that allow for the use of more complex models. Additionally, future work will seek to assess the benefits of the proposed framework in dimensions that extend beyond accuracy. In particular, exploring the benefits for fairness considerations is a core direction for future work. Some experts may be better positioned than others to advocate for members of historically underserved communities, who may also be more likely to be incorrectly classified by the AI tool. In such cases, correctly identifying the expert(s) who can more accurately assess these cases may have important implications for the fairness of decisions resulting from human-AI collaboration.

Finally, human subject studies will be important to determine if quality of decisions can be improved with this type of decision support. In such experiments, the use of the proposed approaches to identify both differing opinions and opinions that align with the AI tool may be explored.

 \subsection*{Acknowledgements}
 This work was in part supported by a Google AI Award for Inclusion Research.

\bibliographystyle{ACM-Reference-Format}
\bibliography{bib}


\begin{thebibliography}{23}

%%% the Web address.

\ifx \showCODEN    \undefined \def \showCODEN     #1{\unskip}     \fi
\ifx \showDOI      \undefined \def \showDOI       #1{#1}\fi
\ifx \showISBNx    \undefined \def \showISBNx     #1{\unskip}     \fi
\ifx \showISBNxiii \undefined \def \showISBNxiii  #1{\unskip}     \fi
\ifx \showISSN     \undefined \def \showISSN      #1{\unskip}     \fi
\ifx \showLCCN     \undefined \def \showLCCN      #1{\unskip}     \fi
\ifx \shownote     \undefined \def \shownote      #1{#1}          \fi
\ifx \showarticletitle \undefined \def \showarticletitle #1{#1}   \fi
\ifx \showURL      \undefined \def \showURL       {\relax}        \fi
% invisible to TeX
\providecommand\bibfield[2]{#2}
\providecommand\bibinfo[2]{#2}
\providecommand\natexlab[1]{#1}
\providecommand\showeprint[2][]{arXiv:#2}

\bibitem[Bansal et~al\mbox{.}(2019)]%
        {bansal2019case}
\bibfield{author}{\bibinfo{person}{Gagan Bansal}, \bibinfo{person}{Besmira
  Nushi}, \bibinfo{person}{Ece Kamar}, \bibinfo{person}{Dan Weld},
  \bibinfo{person}{Walter Lasecki}, {and} \bibinfo{person}{Eric Horvitz}.}
  \bibinfo{year}{2019}\natexlab{}.
\newblock \showarticletitle{A case for backward compatibility for human-ai
  teams}.
\newblock \bibinfo{journal}{\emph{ICML Workshop on Human in the Loop Learning}}
  (\bibinfo{year}{2019}).
\newblock


\bibitem[Bansal et~al\mbox{.}(2021)]%
        {bansal2021does}
\bibfield{author}{\bibinfo{person}{Gagan Bansal}, \bibinfo{person}{Tongshuang
  Wu}, \bibinfo{person}{Joyce Zhou}, \bibinfo{person}{Raymond Fok},
  \bibinfo{person}{Besmira Nushi}, \bibinfo{person}{Ece Kamar},
  \bibinfo{person}{Marco~Tulio Ribeiro}, {and} \bibinfo{person}{Daniel Weld}.}
  \bibinfo{year}{2021}\natexlab{}.
\newblock \showarticletitle{Does the whole exceed its parts? the effect of ai
  explanations on complementary team performance}. In
  \bibinfo{booktitle}{\emph{Proceedings of the CHI Conference on Human Factors
  in Computing Systems}}. \bibinfo{pages}{1--16}.
\newblock


\bibitem[Bhatt et~al\mbox{.}(2020)]%
        {bhatt2020uncertainty}
\bibfield{author}{\bibinfo{person}{Umang Bhatt}, \bibinfo{person}{Javier
  Antor{\'a}n}, \bibinfo{person}{Yunfeng Zhang}, \bibinfo{person}{Q~Vera Liao},
  \bibinfo{person}{Prasanna Sattigeri}, \bibinfo{person}{Riccardo Fogliato},
  \bibinfo{person}{Gabrielle~Gauthier Melan{\c{c}}on},
  \bibinfo{person}{Ranganath Krishnan}, \bibinfo{person}{Jason Stanley},
  \bibinfo{person}{Omesh Tickoo}, {et~al\mbox{.}}}
  \bibinfo{year}{2020}\natexlab{}.
\newblock \showarticletitle{Uncertainty as a form of transparency: Measuring,
  communicating, and using uncertainty}.
\newblock \bibinfo{journal}{\emph{arXiv preprint arXiv:2011.07586}}
  (\bibinfo{year}{2020}).
\newblock


\bibitem[Bu{\c{c}}inca et~al\mbox{.}(2021)]%
        {buccinca2021trust}
\bibfield{author}{\bibinfo{person}{Zana Bu{\c{c}}inca},
  \bibinfo{person}{Maja~Barbara Malaya}, {and} \bibinfo{person}{Krzysztof~Z
  Gajos}.} \bibinfo{year}{2021}\natexlab{}.
\newblock \showarticletitle{To trust or to think: cognitive forcing functions
  can reduce overreliance on AI in AI-assisted decision-making}.
\newblock \bibinfo{journal}{\emph{Proceedings of the ACM on Human-Computer
  Interaction}} \bibinfo{volume}{5}, \bibinfo{number}{CSCW1}
  (\bibinfo{year}{2021}), \bibinfo{pages}{1--21}.
\newblock


\bibitem[Caruana et~al\mbox{.}(2015)]%
        {caruana2015intelligible}
\bibfield{author}{\bibinfo{person}{Rich Caruana}, \bibinfo{person}{Yin Lou},
  \bibinfo{person}{Johannes Gehrke}, \bibinfo{person}{Paul Koch},
  \bibinfo{person}{Marc Sturm}, {and} \bibinfo{person}{Noemie Elhadad}.}
  \bibinfo{year}{2015}\natexlab{}.
\newblock \showarticletitle{Intelligible models for healthcare: Predicting
  pneumonia risk and hospital 30-day readmission}. In
  \bibinfo{booktitle}{\emph{Proceedings of the 21th ACM SIGKDD international
  conference on knowledge discovery and data mining}}.
  \bibinfo{pages}{1721--1730}.
\newblock


\bibitem[Cook(1986)]%
        {cook1986assessment}
\bibfield{author}{\bibinfo{person}{R~Dennis Cook}.}
  \bibinfo{year}{1986}\natexlab{}.
\newblock \showarticletitle{Assessment of local influence}.
\newblock \bibinfo{journal}{\emph{Journal of the Royal Statistical Society:
  Series B (Methodological)}} \bibinfo{volume}{48}, \bibinfo{number}{2}
  (\bibinfo{year}{1986}), \bibinfo{pages}{133--155}.
\newblock


\bibitem[De-Arteaga et~al\mbox{.}(2021)]%
        {de2021leveraging}
\bibfield{author}{\bibinfo{person}{Maria De-Arteaga}, \bibinfo{person}{Artur
  Dubrawski}, {and} \bibinfo{person}{Alexandra Chouldechova}.}
  \bibinfo{year}{2021}\natexlab{}.
\newblock \showarticletitle{Leveraging Expert Consistency to Improve
  Algorithmic Decision Support}.
\newblock \bibinfo{journal}{\emph{arXiv preprint arXiv:2101.09648}}
  (\bibinfo{year}{2021}).
\newblock


\bibitem[De-Arteaga et~al\mbox{.}(2020)]%
        {de2020case}
\bibfield{author}{\bibinfo{person}{Maria De-Arteaga}, \bibinfo{person}{Riccardo
  Fogliato}, {and} \bibinfo{person}{Alexandra Chouldechova}.}
  \bibinfo{year}{2020}\natexlab{}.
\newblock \showarticletitle{A Case for Humans-in-the-Loop: Decisions in the
  Presence of Erroneous Algorithmic Scores}. In
  \bibinfo{booktitle}{\emph{Proceedings of the 2020 CHI Conference on Human
  Factors in Computing Systems}}. \bibinfo{pages}{1--12}.
\newblock


\bibitem[Dietvorst et~al\mbox{.}(2015)]%
        {dietvorst2015algorithm}
\bibfield{author}{\bibinfo{person}{Berkeley~J Dietvorst},
  \bibinfo{person}{Joseph~P Simmons}, {and} \bibinfo{person}{Cade Massey}.}
  \bibinfo{year}{2015}\natexlab{}.
\newblock \showarticletitle{Algorithm aversion: People erroneously avoid
  algorithms after seeing them err.}
\newblock \bibinfo{journal}{\emph{Journal of Experimental Psychology: General}}
  \bibinfo{volume}{144}, \bibinfo{number}{1} (\bibinfo{year}{2015}),
  \bibinfo{pages}{114}.
\newblock


\bibitem[Fernandes et~al\mbox{.}(2017)]%
        {fernandes2017transfer}
\bibfield{author}{\bibinfo{person}{Kelwin Fernandes}, \bibinfo{person}{Jaime~S
  Cardoso}, {and} \bibinfo{person}{Jessica Fernandes}.}
  \bibinfo{year}{2017}\natexlab{}.
\newblock \showarticletitle{Transfer learning with partial observability
  applied to cervical cancer screening}. In \bibinfo{booktitle}{\emph{Iberian
  conference on pattern recognition and image analysis}}. Springer,
  \bibinfo{pages}{243--250}.
\newblock


\bibitem[Gal and Ghahramani(2016)]%
        {gal2016dropout}
\bibfield{author}{\bibinfo{person}{Yarin Gal} {and} \bibinfo{person}{Zoubin
  Ghahramani}.} \bibinfo{year}{2016}\natexlab{}.
\newblock \showarticletitle{Dropout as a bayesian approximation: Representing
  model uncertainty in deep learning}. In
  \bibinfo{booktitle}{\emph{International Conference on Machine Learning
  (ICML)}}. \bibinfo{pages}{1050--1059}.
\newblock


\bibitem[Gao et~al\mbox{.}(2021)]%
        {gao2021human}
\bibfield{author}{\bibinfo{person}{Ruijiang Gao}, \bibinfo{person}{Maytal
  Saar-Tsechansky}, \bibinfo{person}{Maria De-Arteaga}, \bibinfo{person}{Ligong
  Han}, \bibinfo{person}{Min~Kyung Lee}, {and} \bibinfo{person}{Matthew
  Lease}.} \bibinfo{year}{2021}\natexlab{}.
\newblock \showarticletitle{Human-{AI} Collaboration with Bandit Feedback}.
\newblock \bibinfo{journal}{\emph{IJCAI}} (\bibinfo{year}{2021}).
\newblock


\bibitem[Koh and Liang(2017)]%
        {koh2017understanding}
\bibfield{author}{\bibinfo{person}{Pang~Wei Koh} {and} \bibinfo{person}{Percy
  Liang}.} \bibinfo{year}{2017}\natexlab{}.
\newblock \showarticletitle{Understanding black-box predictions via influence
  functions}. In \bibinfo{booktitle}{\emph{Proceedings of the 34th
  International Conference on Machine Learning-Volume 70}}. JMLR. org,
  \bibinfo{pages}{1885--1894}.
\newblock


\bibitem[Lakkaraju et~al\mbox{.}(2017)]%
        {lakkaraju2017selective}
\bibfield{author}{\bibinfo{person}{Himabindu Lakkaraju}, \bibinfo{person}{Jon
  Kleinberg}, \bibinfo{person}{Jure Leskovec}, \bibinfo{person}{Jens Ludwig},
  {and} \bibinfo{person}{Sendhil Mullainathan}.}
  \bibinfo{year}{2017}\natexlab{}.
\newblock \showarticletitle{The Selective Labels Problem: Evaluating
  Algorithmic Predictions in the Presence of Unobservables}. In
  \bibinfo{booktitle}{\emph{Proceedings of the 23rd ACM SIGKDD International
  Conference on Knowledge Discovery and Data Mining}}. ACM,
  \bibinfo{pages}{275--284}.
\newblock


\bibitem[Lebovitz et~al\mbox{.}(2020)]%
        {lebovitz2020incorporate}
\bibfield{author}{\bibinfo{person}{Sarah Lebovitz}, \bibinfo{person}{Hila
  Lifshitz-Assaf}, {and} \bibinfo{person}{Natalia Levina}.}
  \bibinfo{year}{2020}\natexlab{}.
\newblock \showarticletitle{To incorporate or not to incorporate AI for
  critical judgments: The importance of ambiguity in professionals’ judgment
  process}.
\newblock \bibinfo{journal}{\emph{NYU Stern School of Business}}
  (\bibinfo{year}{2020}).
\newblock


\bibitem[Maddox et~al\mbox{.}(2019)]%
        {maddox2019simple}
\bibfield{author}{\bibinfo{person}{Wesley~J Maddox}, \bibinfo{person}{Pavel
  Izmailov}, \bibinfo{person}{Timur Garipov}, \bibinfo{person}{Dmitry~P
  Vetrov}, {and} \bibinfo{person}{Andrew~Gordon Wilson}.}
  \bibinfo{year}{2019}\natexlab{}.
\newblock \showarticletitle{A simple baseline for bayesian uncertainty in deep
  learning}.
\newblock \bibinfo{journal}{\emph{Advances in Neural Information Processing
  Systems}}  \bibinfo{volume}{32} (\bibinfo{year}{2019}),
  \bibinfo{pages}{13153--13164}.
\newblock


\bibitem[Madras et~al\mbox{.}(2018)]%
        {madras2018predict}
\bibfield{author}{\bibinfo{person}{David Madras}, \bibinfo{person}{Toni
  Pitassi}, {and} \bibinfo{person}{Richard Zemel}.}
  \bibinfo{year}{2018}\natexlab{}.
\newblock \showarticletitle{Predict responsibly: improving fairness and
  accuracy by learning to defer}.
\newblock \bibinfo{journal}{\emph{NeurIPS}}  \bibinfo{volume}{31}
  (\bibinfo{year}{2018}), \bibinfo{pages}{6147--6157}.
\newblock


\bibitem[McGrath et~al\mbox{.}(2020)]%
        {mcgrath2020does}
\bibfield{author}{\bibinfo{person}{Sean McGrath}, \bibinfo{person}{Parth
  Mehta}, \bibinfo{person}{Alexandra Zytek}, \bibinfo{person}{Isaac Lage},
  {and} \bibinfo{person}{Himabindu Lakkaraju}.}
  \bibinfo{year}{2020}\natexlab{}.
\newblock \showarticletitle{When Does Uncertainty Matter?: Understanding the
  Impact of Predictive Uncertainty in ML Assisted Decision Making}.
\newblock \bibinfo{journal}{\emph{arXiv preprint arXiv:2011.06167}}
  (\bibinfo{year}{2020}).
\newblock


\bibitem[Raghu et~al\mbox{.}(2019)]%
        {raghu2018direct}
\bibfield{author}{\bibinfo{person}{Maithra Raghu}, \bibinfo{person}{Katy
  Blumer}, \bibinfo{person}{Rory Sayres}, \bibinfo{person}{Ziad Obermeyer},
  \bibinfo{person}{Robert Kleinberg}, \bibinfo{person}{Sendhil Mullainathan},
  {and} \bibinfo{person}{Jon Kleinberg}.} \bibinfo{year}{2019}\natexlab{}.
\newblock \showarticletitle{Direct uncertainty prediction for medical second
  opinions}.
\newblock \bibinfo{journal}{\emph{International Conference on Machine Learning
  (ICML)}} (\bibinfo{year}{2019}).
\newblock


\bibitem[Ribeiro et~al\mbox{.}(2016)]%
        {ribeiro2016should}
\bibfield{author}{\bibinfo{person}{Marco~Tulio Ribeiro},
  \bibinfo{person}{Sameer Singh}, {and} \bibinfo{person}{Carlos Guestrin}.}
  \bibinfo{year}{2016}\natexlab{}.
\newblock \showarticletitle{" Why should i trust you?" Explaining the
  predictions of any classifier}. In \bibinfo{booktitle}{\emph{Proceedings of
  the 22nd ACM SIGKDD international conference on knowledge discovery and data
  mining}}. \bibinfo{pages}{1135--1144}.
\newblock


\bibitem[Skitka et~al\mbox{.}(2000)]%
        {skitka2000accountability}
\bibfield{author}{\bibinfo{person}{Linda~J Skitka}, \bibinfo{person}{Kathleen
  Mosier}, {and} \bibinfo{person}{Mark~D Burdick}.}
  \bibinfo{year}{2000}\natexlab{}.
\newblock \showarticletitle{Accountability and automation bias}.
\newblock \bibinfo{journal}{\emph{International Journal of Human-Computer
  Studies}} \bibinfo{volume}{52}, \bibinfo{number}{4} (\bibinfo{year}{2000}),
  \bibinfo{pages}{701--717}.
\newblock


\bibitem[Wilder et~al\mbox{.}(2020)]%
        {wilder2020learning}
\bibfield{author}{\bibinfo{person}{Bryan Wilder}, \bibinfo{person}{Eric
  Horvitz}, {and} \bibinfo{person}{Ece Kamar}.}
  \bibinfo{year}{2020}\natexlab{}.
\newblock \showarticletitle{Learning to Complement Humans}.
\newblock \bibinfo{journal}{\emph{IJCAI}} (\bibinfo{year}{2020}).
\newblock


\bibitem[Zhang et~al\mbox{.}(2020)]%
        {zhang2020effect}
\bibfield{author}{\bibinfo{person}{Yunfeng Zhang}, \bibinfo{person}{Q~Vera
  Liao}, {and} \bibinfo{person}{Rachel~KE Bellamy}.}
  \bibinfo{year}{2020}\natexlab{}.
\newblock \showarticletitle{Effect of confidence and explanation on accuracy
  and trust calibration in ai-assisted decision making}. In
  \bibinfo{booktitle}{\emph{Proceedings of the 2020 Conference on Fairness,
  Accountability, and Transparency}}. \bibinfo{pages}{295--305}.
\newblock


\end{thebibliography}

\end{document}